\documentclass{article} 
\usepackage{iclr2027_conference,times}
 \iclrfinalcopy

\usepackage{multirow}
\usepackage{amsmath, amssymb, amsthm}
\usepackage{graphicx}
\usepackage{makecell}

\usepackage{amsmath,amsfonts,bm}

\def\eqref#1{equation~\ref{#1}}

\def\1{\bm{1}}

\DeclareMathAlphabet{\mathsfit}{\encodingdefault}{\sfdefault}{m}{sl}
\SetMathAlphabet{\mathsfit}{bold}{\encodingdefault}{\sfdefault}{bx}{n}

\usepackage{hyperref}
\usepackage{url}
\usepackage[utf8]{inputenc} 
\usepackage[T1]{fontenc}    
\usepackage{hyperref}       
\usepackage{url}            
\usepackage{booktabs}       
\usepackage{amsfonts}       
\usepackage{nicefrac}       
\usepackage{microtype}      
\usepackage{xcolor} 

\title{Nested Inductive Bias Framework for SPD Manifold Learning}

\author{Tushar Das  \\
National Institute of Technology Jamshedpur\\
\texttt{tdas2663@gmail.com} \\
}

\begin{document}

\maketitle
\begin{abstract}
In Geometric Deep Learning, inductive biases serve two primary functions: enforcing manifold constraints and embedding relational priors. Currently, representation learning on SPD manifolds frequently relies on pullback Euclidean metrics, such as the Log-Euclidean Metric, to satisfy the former. While computationally efficient in avoiding domain boundary violations, these metrics induce a flat geometry that may fail to capture the intrinsic relational priors of datasets. While metrics such as the Poincaré metric are widely utilized to induce domain-aligned relational priors, generalizing them from standard vector representations to the SPD manifold has remained a challenge. To bridge this gap, we introduce a Nested Inductive Bias framework that utilizes a two-stage diffeomorphic composition to formally pull back non-Euclidean target geometries onto the SPD manifold. This framework enables the construction of curvature-aligned Riemannian classifiers that simultaneously respect matrix constraints and the latent relational geometry of the data. Empirical evaluations on kinematic and signal processing benchmarks, together with synthetic experiments, demonstrate that deep manifold networks experience degradation in class separability unless the metric curvature aligns with the intrinsic data distribution. Furthermore, for standard vectorized architectures, we propose the Rational Conformal Metric (RCM), designed to establish state-of-the-art geometric robustness against outliers by bounding the representation space.
\end{abstract}

\section{Introduction}
\label{sec:intro}
Symmetric Positive Definite (SPD) matrices have emerged as fundamental mathematical representations across different domains, enabling significant progress in medical image analysis \citep{chakraborty2020manifoldnet}, neural signal decoding \citep{kobler2022spd, li2025spdim}, kinematic modeling \citep{huang2017riemannian}, and radio-frequency target identification \citep{brooks2019riemannian}. Geometrically, these matrices reside on a strictly positive, convex conical manifold. Applying standard Euclidean computations directly to these representations violates the manifold boundary. To operate within these manifold boundaries, the field of Riemannian deep learning relies on metric tensors, such as the Affine-Invariant Riemannian Metric (AIRM) \citep{pennec2006riemannian} and Pullback Euclidean metrics like the Log-Euclidean Metric (LEM) \citep{arsigny2006log}. These metrics have enabled the adaptation of dense neural network components, such as residual connections \citep{katsman2024riemannian}, normalization layers \citep{brooks2019riemannian}, and universal parameterized metrics \citep{das2026beyond}, for manifold-valued data. Furthermore, the theoretical formalization of \textit{deformed metrics} \citep{thanwerdas2022geometry} has demonstrated that valid Riemannian structures can be systematically generated via diffeomorphic pullbacks.

However, this classical paradigm reveals a gap in Geometric Deep Learning: the disconnect between enforcing \textit{domain constraints} and embedding \textit{relational priors}. For vector data (such as natural language), researchers often inject relational priors, such as Hyperbolic geometry \citep{ganea2018hyperbolic} and the unified $\kappa$-stereographic model \citep{skopek2020mixed}, to exploit hierarchical or spherical structures. Conversely, for structured data like SPD matrices, the manifold is often treated purely as a physical constraint. In recent literature, pullback Euclidean metrics, like the Log-Euclidean Metric, have gained popularity because their logarithmic maps safely project SPD matrices into a Euclidean tangent space for unconstrained deep learning operations \citep{arsigny2006log, huang2017riemannian,yger2017riemannian}. This prevents the violation of the SPD constraint, but effectively stops at a "Level 1" inductive bias. We argue that satisfying this domain constraint is necessary, but geometrically insufficient. Real-world SPD datasets frequently possess secondary relational biases, such as the kinematic hierarchies inherent in human action data. Forcing these tree-like distributions into the flat geometry of LEM may ignore their latent data geometry, leading to degraded model performance. Besides, the popular, non-positive-curvature-inducing AIRM may not be suitable for non-hierarchical datasets and may incur substantial computational overhead. 

To bridge this limitation, we introduce the framework of \textbf{Nested Inductive Biases}. We postulate that optimal representation learning on matrix manifolds requires simultaneously satisfying the physical constraints of the data and its latent relational structure. We achieve this via a two-stage diffeomorphic pullback. First, SPD matrices are mapped to the Log-Euclidean tangent space at the identity, thereby satisfying the domain constraint. Second, we identify this intermediate Euclidean vector space with the origin-centered tangent space of a secondary target manifold (such as the Poincaré ball or the Projected Hypersphere). By applying the Riemannian exponential map of the target, we map these coordinates onto the target manifold. Composing these mappings allows us to pull the target metrics back onto the SPD manifold, endowing it with computationally efficient Hyperbolic and Spherical inductive biases for the first time. This enables the utilization of the native RMLR \citep{chen2024rmlr} for classification without relying on tangent-space Euclidean classification. Furthermore, we introduce the Rational Conformal Metric, a novel bounded geometry designed to provide regularization against outliers.

The main contributions of our work are summarized as follows:
\begin{itemize}
    \item We formalize the theory of Nested Inductive Biases, defining a two-stage diffeomorphic pullback mechanism that bridges matrix manifold constraints with non-Euclidean relational priors traditionally limited to vector embeddings.
    \item We derive computationally efficient Riemannian Multinomial Logistic Regression (RMLR) layers for the Nested Hyperbolic Metric (NHM), which induces a negative curvature without AIRM's computational overhead, and Nested Spherical Metric (NSM), alongside a Tangent Space Mapping (TSM) framework for the geodesically incomplete Rational Conformal Metric (RCM).
    \item We empirically validate our framework across kinematic and signal processing datasets using Linear Probes, SPDNet, and deep RResNet architectures. We demonstrate that aligning the geometric prior with the intrinsic data structures yields superior performance, corroborated through controlled synthetic geometric ablations.
\end{itemize}

\section{Preliminaries:The $\kappa$-Stereographic Model}
To unify target geometries of varying constant sectional curvatures $K$, we adopt the generalized $\kappa$-stereographic model \citep{skopek2020mixed}, where $\kappa = K$ and $c = \vert{}K\vert{}$. This transitions between the Poincaré ball ($\kappa < 0$), Euclidean space ($\kappa = 0$), and the Projected Hypersphere ($\kappa > 0$). The manifold is defined as $\mathcal{M}_\kappa^n = \{z \in \mathbb{R}^n : -\kappa \|z\|^2 < 1\}$.

The geometry is defined by the conformal factor $\lambda_z^\kappa = \frac{2}{1 + \kappa \|z\|^2}$, yielding the metric tensor $g_z^\kappa = (\lambda_z^\kappa)^2 \mathbf{I}$. For operations mapping between the origin-centered tangent space $\mathcal{T}_{\mathbf{0}}\mathcal{M}_\kappa^n$ and the manifold, the exponential and logarithmic maps are defined using generalized trigonometric functions:
\begin{equation}
    \exp_{\mathbf{0}}^\kappa(v) = \tan_\kappa\left( \sqrt{|\kappa|} \|v\| \right) \frac{v}{\sqrt{|\kappa|} \|v\|}
\end{equation}
\begin{equation}
    \log_{\mathbf{0}}^\kappa(z) = \tan_\kappa^{-1}\left( \sqrt{|\kappa|} \|z\| \right) \frac{z}{\sqrt{|\kappa|} \|z\|}
\end{equation}
where $\tan_\kappa$ and $\tan_\kappa^{-1}$ correspond to $\tanh$ and $\text{arctanh}$ for $\kappa < 0$, and $\tan$ and $\arctan$ for $\kappa > 0$. 

\section{The Nested Inductive Biases Framework}
\label{sec:theory}

An open and convex cone within the ambient space of symmetric matrices, the Symmetric Positive Definite (SPD) matrix manifold, $\mathcal{S}_{++}^d$, possesses no inherent geometric structure or curvature. A geometry is only realized when the manifold is endowed with a Riemannian metric tensor. Classical formulations typically endow $\mathcal{S}_{++}^d$ with the Affine-Invariant Riemannian Metric (AIRM), which induces non-positive sectional curvature but requires computationally intensive $\mathcal{O}(d^3)$ eigendecompositions and iterative approximations for basic operations. In recent years, pullback Euclidean metrics such as the Log-Euclidean Metric (LEM) have gained popularity, offering computational efficiency but inducing flat ($K=0$) curvature, thereby precluding the ability to model curved relational structures.

To resolve this dichotomy, we introduce the framework of Nested Inductive Biases. By establishing a two-stage diffeomorphic composition, we execute an isometric pullback of target non-Euclidean geometries—specifically the Poincaré ball $\mathbb{P}_c^n$ and the Hypersphere $\mathbb{D}_c^n$—directly onto the SPD cone. This endows the SPD manifold with constant non-zero sectional curvature while providing the computational efficiency akin to pullback Euclidean metrics. Furthermore, unlike applying the same metric to different types of datasets, we show the importance of dataset-aligned metric use.

\subsection{The Two-Stage Diffeomorphic Composition}
\label{subsec:diffeo_composition}

We construct a composite mapping $\Phi : \mathcal{S}_{++}^d \to \mathcal{M}_{target}$ (where $\mathcal{M}_{target}$ represents a manifold possessing constant sectional curvature $K \neq 0$). This is formulated via $\Phi = \phi_2 \circ \phi_1$.

\textbf{Proposition 1 } \textit{Let $\phi_1 : \mathcal{S}_{++}^d \to \mathbb{R}^n$ (where $n = d(d+1)/2$) be defined by mapping an SPD matrix $\mathbf{S}$ to the tangent space at the identity $\mathcal{T}_{\mathbf{I}}\mathcal{S}_{++}^d$ via the principal matrix logarithm, followed by the half-vectorization operator $\text{vech}(\cdot)$:}
\begin{equation}
    \mathbf{v} = \phi_1(\mathbf{S}) = \text{vech}(\text{Log}_{\mathbf{I}}(\mathbf{S}))
\end{equation}
\textit{The mapping $\phi_1$ is a global diffeomorphism.}

\textbf{Proof.} The principal matrix logarithm $\text{Log}_{\mathbf{I}} : \mathcal{S}_{++}^d \to \mathbb{S}^d$ is a well-established global $C^\infty$ diffeomorphism between the SPD cone and the vector space of symmetric matrices $\mathbb{S}^d$. The half-vectorization operator $\text{vech} : \mathbb{S}^d \to \mathbb{R}^n$ (where $n = d(d+1)/2$) is a bijective linear transformation. Because all finite-dimensional linear isomorphisms are globally $C^\infty$ smooth with smooth inverses, $\text{vech}$ acts as a global diffeomorphism between $\mathbb{S}^d$ and $\mathbb{R}^n$. The composition of two diffeomorphisms mathematically guarantees that $\phi_1$ is a global diffeomorphism from $\mathcal{S}_{++}^d$ to $\mathbb{R}^n$. \hfill $\blacksquare$

\textbf{Proposition 2} \textit{Let $\phi_2 : \mathbb{R}^n \to \mathcal{M}_{target}$ be defined by mapping the Euclidean vector $\mathbf{v}$ onto the target manifold via the Riemannian exponential map at the origin:}
\begin{equation}
    \mathbf{z} = \phi_2(\mathbf{v}) = \exp_{\mathbf{0}}^{\mathcal{M}_{target}}(\mathbf{v})
\end{equation}
\textit{Let $c = \vert{}K\vert{}$ denote the absolute constant curvature of the target manifold. For the Poincaré ball $\mathbb{P}_c^n$, the exponential map is a global diffeomorphism. For the Projected Hypersphere $\mathbb{D}_c^n$, $\phi_2$ is a diffeomorphism within the injectivity radius $\pi/\sqrt{c}$.}

\textbf{Proof.} By the Cartan-Hadamard theorem, any simply connected, complete Riemannian manifold with non-positive sectional curvature (such as the Poincaré ball $\mathbb{P}_c^n$) contains no conjugate points. Consequently, its Riemannian exponential map $\exp_{\mathbf{0}}$ is a global diffeomorphism from the tangent space to the manifold. For the Projected Hypersphere $\mathbb{D}_c^n$ ($K > 0$), conjugate points (antipodes) exist at a distance of $\pi/\sqrt{c}$. Therefore, the exponential map remains a strict diffeomorphism provided $\|\mathbf{v}\| < \pi/\sqrt{c}$, avoiding the cut locus. \hfill $\blacksquare$

\textbf{Theorem 1 (Nested Diffeomorphism).} \textit{The nested mapping $\Phi(\mathbf{S}) = \phi_2(\phi_1(\mathbf{S}))$ constitutes a diffeomorphism between the SPD manifold and the target geometry $\mathcal{M}_{target}$ (globally for $K \le 0$, and within the injectivity radius $\pi/\sqrt{c}$ for $K>0$).}

\textbf{Proof.} By the Inverse Function Theorem, the composition of two $C^\infty$ diffeomorphisms yields a $C^\infty$ diffeomorphism. Since $\phi_1$ is globally diffeomorphic and $\phi_2$ is diffeomorphic within its respective bounds, their composition $\Phi = \phi_2 \circ \phi_1$, for the Poincaré ball, provides a smooth bijection with a smooth inverse, $\Phi^{-1} = \phi_1^{-1} \circ \phi_2^{-1}$, globally, and so within the injectivity radius for the Projected Hypersphere. \hfill $\blacksquare$

\subsection{Induced Curvature via Isometric Pullback}
\label{subsec:isometric_pullback}

\textbf{Theorem 2} \textit{Let $g_{target}$ denote the Riemannian metric tensor of $\mathcal{M}_{target}$. We define the nested Riemannian metric $g_{nested}$ on $\mathcal{S}_{++}^d$ as the pullback of $g_{target}$ by $\Phi$:}
\begin{equation}
    g_{nested} = \Phi^* g_{target}
\end{equation}
\textit{For any $\mathbf{S} \in \mathcal{S}_{++}^d$ and tangent vectors $\mathbf{V}, \mathbf{W} \in \mathcal{T}_{\mathbf{S}}\mathcal{S}_{++}^d$, the inner product evaluates to:}
\begin{equation}
    (g_{nested})_{\mathbf{S}}(\mathbf{V}, \mathbf{W}) = (g_{target})_{\Phi(\mathbf{S})} \left( \Phi_{*, \mathbf{S}}(\mathbf{V}), \Phi_{*, \mathbf{S}}(\mathbf{W}) \right)
\end{equation}
\textit{where $\Phi_{*, \mathbf{S}}$ is the pushforward differential of $\Phi$. Under this metric, $\Phi$ acts as an isometry between $(\mathcal{S}_{++}^d, g_{nested})$ and $(\mathcal{M}_{target}, g_{target})$, ensuring that the SPD manifold strictly inherits the constant sectional curvature of the target space.}

\textbf{Proof.} By definition of the pullback operation, equipping a manifold with a pullback metric via a diffeomorphism preserves the first fundamental form, rendering the mapping an isometry. By the Fundamental Theorem of Riemannian Geometry, the metric tensor uniquely determines the Levi-Civita connection and the associated Riemann curvature tensor. Because local isometries preserve the metric tensor, the sectional curvature of the SPD manifold under $g_{nested}$ is equal to the target manifold's curvature at all points. \hfill $\blacksquare$

\subsection{Nested Riemannian Multinomial Logistic Regression (Track A)}
\label{subsec:nested_rmlr}

Because the SPD manifold equipped with $g_{nested}$ now possesses constant non-zero curvature, linear classification cannot be performed via standard Euclidean hyperplanes. We formulate the classification boundary directly within the target geometry using generalized Riemannian Multinomial Logistic Regression (RMLR) \citep{chen2024rmlr}.

\textbf{Definition 1 (Nested Riemannian Hyperplane).} \textit{For a target class $k$, let $\mathbf{P}_k \in \mathcal{S}_{++}^d$ denote the learned class anchor, and $\mathbf{A}_k \in \mathcal{T}_{\mathbf{P}_k}\mathcal{S}_{++}^d \setminus \{\mathbf{0}\}$ denote a normal tangent vector. Let $p_k = \Phi(\mathbf{P}_k) \in \mathcal{M}_{target}$ and $a_k = \Phi_{*, \mathbf{P}_k}(\mathbf{A}_k)$. Using the Riemannian Logarithmic map, the decision boundary $\tilde{H}_{a_k, p_k}$ is defined by the locus of points whose projected tangent vector is orthogonal to the normal vector:}
\begin{equation}
    \tilde{H}_{a_k, p_k} = \left\{ \mathbf{S} \in \mathcal{S}_{++}^d : \langle \text{Log}_{p_k}(\Phi(\mathbf{S})), a_k \rangle_{p_k} = 0 \right\}
\end{equation}

\textbf{Theorem 3 (Nested RMLR Classification).} \textit{Following the generalized RMLR framework, the multinomial probability for class $k$ evaluates directly via the Riemannian inner product within the tangent space of the class centroid:}
\begin{equation}
    p(y=k | \mathbf{S}) \propto \exp \left( \langle \text{Log}_{p_k}(\Phi(\mathbf{S})), a_k \rangle_{p_k} \right)
\end{equation}
\textit{where $a_k \in \mathcal{T}_{\mathbf{0}}\mathcal{M}_{target} \setminus \{\mathbf{0}\}$ is optimized as a Euclidean parameter within the origin tangent space $\mathcal{T}_{\mathbf{0}}\mathcal{M}_{target}$ and parallel transported to $p_k$.}

\textbf{Proof.} Let $\mathbf{z} = \Phi(\mathbf{S}) \in \mathcal{M}_{target}$ be the mapped point. By the definition of the Riemannian inner product, we have $\langle \text{Log}_{p_k}(z), a_k \rangle_{p_k} = \|\text{Log}_{p_k}(z)\|_{p_k} \|a_k\|_{p_k} \cos(\theta)$, where $\theta$ is the angle between the tangent vectors. Following the generalized RMLR framework \citet{chen2024rmlr}, this tangent-space projection encodes the signed orthogonal distance to the decision boundary relative to the class anchor $p_k$. Therefore, the inner product functions as a linear logit for cross-entropy optimization. \hfill $\blacksquare$

\subsection{The Rational Conformal Metric (RCM)}
\label{subsec:rational_metric_def}

While Hyperbolic and Spherical spaces induce non-Euclidean relational priors, they are geodesically complete geometries that do not bound the magnitude of tangent vectors. In real-world sensor domains (such as Radar signal processing), amplitude anomalies may manifest as outliers within the base Log-Euclidean tangent space. To suppress unbounded Euclidean noise on the SPD manifold, we propose a novel bounded metric: the Rational Conformal Metric (RCM).

\textbf{Definition 2 (Rational Conformal Metric).} \textit{We define the Rational Conformal Metric on the Euclidean vector space $\mathbb{R}^n$ via a radially symmetric conformal transformation. For a scale parameter $\alpha > 0$, the conformal factor $\lambda(\mathbf{x})$ that scales infinitesimal lengths is defined as:}
\begin{equation}
    \lambda(\mathbf{x}) = \frac{1}{(1 + \alpha\|\mathbf{x}\|)^2}
\end{equation}
\textit{Consequently, the Riemannian metric tensor is defined as $g^{RCM}_{\mathbf{x}} = \lambda(\mathbf{x})^2 \mathbf{I}$, yielding:}
\begin{equation}
    g^{RCM}_{\mathbf{x}} = \frac{1}{(1 + \alpha\|\mathbf{x}\|)^4} \mathbf{I}
\end{equation}
\textit{where $\mathbf{I}$ is the Euclidean identity matrix. The corresponding conformal factor is decreasing along the radial axis, ensuring that the infinitesimal length element shrinks rationally as points diverge toward infinity.}

By pulling $g^{RCM}$ back onto the SPD cone via our nested framework, we endow the manifold with a geometry designed specifically to compress extreme deviations without altering the angular geometry of the central data distribution.

\subsection{Tangent Space Mapping for Geodesically Incomplete Spaces (Track B)}
\label{subsec:tsm_track_b}

Unlike standard spaces of constant curvature, the Rational Conformal Metric yields a geodesically incomplete manifold. A geodesic ray extending to infinite Euclidean distance possesses a finite Riemannian length. We leverage this deliberate incompleteness to formulate a bounded Tangent Space Mapping (TSM) framework for linear classification.

\textbf{Theorem 4} \textit{Under the Rational Conformal Metric, the origin-centered logarithmic map $\text{Log}_{\mathbf{0}}$ acts as a global diffeomorphism from the unbounded vector space $\mathbb{R}^n$ onto the open Euclidean ball $\mathcal{B}_{1/\alpha}$. Consequently, TSM guarantees a unique and bounded representation for any $\mathbf{x} \in \mathbb{R}^n$, while the domain of inverse exponential map $\text{Exp}_{\mathbf{0}}$ is restricted to the open ball.}

\textbf{Proof.} Under the radially symmetric conformal metric, the geodesic distance from the origin is defined by the integral of the square root of the metric tensor along the radial path:
\begin{equation}
    d_{RCM}(\mathbf{0}, \mathbf{x}) = \int_0^{\|\mathbf{x}\|} \frac{1}{(1 + \alpha r)^2} dr = \frac{1}{\alpha} \left( 1 - \frac{1}{1 + \alpha \|\mathbf{x}\|} \right) = \frac{\|\mathbf{x}\|}{1 + \alpha\|\mathbf{x}\|}
\end{equation}

\begin{equation}
    \mathbf{v} = \text{Log}_{\mathbf{0}}(\mathbf{x}) = d_{RCM}(\mathbf{0}, \mathbf{x}) \frac{\mathbf{x}}{\|\mathbf{x}\|} = \frac{\mathbf{x}}{1 + \alpha\|\mathbf{x}\|}
\end{equation}
Taking the norm of the tangent vector $\mathbf{v}$, we observe the asymptotic bound: $\|\mathbf{v}\| = \frac{\|\mathbf{x}\|}{1 + \alpha\|\mathbf{x}\|} < \frac{1}{\alpha}$ for all $\mathbf{x} \in \mathbb{R}^n$. Thus, $\text{Log}_{\mathbf{0}}$ injectively maps the unbounded vector space into the open ball $\mathcal{B}_{1/\alpha}$. 

To prove bijectivity onto $\mathcal{B}_{1/\alpha}$, we construct the Riemannian Exponential map $\text{Exp}_{\mathbf{0}}(\mathbf{v})$. Given $\mathbf{v} \in \mathcal{B}_{1/\alpha}$, we solve for $\|\mathbf{x}\|$:
\begin{equation}
    \|\mathbf{v}\| = \frac{\|\mathbf{x}\|}{1 + \alpha\|\mathbf{x}\|} \implies \|\mathbf{x}\| = \frac{\|\mathbf{v}\|}{1 - \alpha\|\mathbf{v}\|}
\end{equation}
Substituting $\|\mathbf{x}\|$ back yields the exact closed-form Exponential map: $\mathbf{x} = \text{Exp}_{\mathbf{0}}(\mathbf{v}) = \frac{\mathbf{v}}{1 - \alpha\|\mathbf{v}\|}$. This inverse mapping is uniquely defined and continuous if and only if the denominator is strictly positive ($\|\mathbf{v}\| < 1/\alpha$), formally reflecting the manifold's geodesic incompleteness. By performing linear classification strictly within the mapped tangent space (TSM), the optimization boundary mathematically ignores the missing inverse projection, ensuring stability while physically compressing unbounded vectors into $\mathcal{B}_{1/\alpha}$. \hfill $\blacksquare$

\textbf{Remark 2 (Computational and Numerical Efficiency).} \textit{Beyond its topological utility for outlier compression, the Rational Conformal Metric offers a practical computational advantage for deep learning. Standard constant-curvature manifolds (such as the $\kappa$-stereographic models) rely heavily on transcendental functions ($\tan$, $\arctan$, $\tanh$) and square roots to compute their exponential and logarithmic maps. In contrast, the RCM's $\text{Exp}_{\mathbf{0}}$ and $\text{Log}_{\mathbf{0}}$ mappings evaluate entirely via basic rational polynomials, eliminating the numerical instabilities commonly associated with trigonometric backpropagation in automatic differentiation frameworks.}

\section{Experimental Methodology}
\label{sec:methodology}

While standard RMLR formulations evaluate class probabilities using standard metrics (such as AIRM or LEM), our nested mechanism allows us to impose custom geometric priors. By composing the Log-Euclidean mapping with the $\kappa$-stereographic exponential map (as derived before), we induce a family of composite Riemannian metrics on the SPD manifold. We denote these as \textbf{NHM} (Nested Hyperbolic Metric, $K < 0$) and \textbf{NSM} (Nested Spherical Metric, $K > 0$). This framework allows the subsequent RMLR layer to compute decision boundaries that simultaneously respect the symmetric positive definiteness of the data and the intrinsic relational curvature of the dataset.

\subsection{Datasets and Preprocessing}
\label{subsec:datasets}

Empirical validation is conducted on two popular datasets to assess representation learning on SPD manifolds:
\begin{itemize}
    \item \textbf{First-Person Hand Action (FPHA) \citep{garcia2018first}:} Representing human kinematics, this task utilizes 3D joint trajectories converted into temporal covariance matrices ($63 \times 63$ dimensions). Performance is reported on the official 45-class action recognition test split.
    \item \textbf{Radar Target Classification \citep{brooks2019riemannian}:} A signal processing benchmark consisting of $20 \times 20$ covariance matrices captured under varying signal-to-noise conditions. Generalization is evaluated via 5-fold cross-validation.
\end{itemize}

To isolate metric inductive bias from network capacity, we evaluate three architectures::

\textbf{1. Linear Probe:} Maps inputs directly to the classification space, isolating the linear separability induced solely by the metric mapping.

\textbf{2. SPDNet:} Incorporates sequential Bilinear Mapping (BiMap) and Rectified Eigenvalue (ReEig) layers \citep{huang2017riemannian}. To prevent confounding factors from information bottlenecking, manifold dimensions are strictly preserved throughout the forward pass ($[63 \to 63]$ for FPHA, $[20 \to 20]$ for Radar).

\textbf{3. Deep RResNet:} A deep Riemannian Residual architecture \citep{katsman2024riemannian} incorporating manifold-constrained skip connections. This model evaluates metric robustness in compressed latent spaces by applying a dimensionality-reducing BiMap before the residual blocks ($[63 \to 33]$ for FPHA, $[20 \to 8]$ for Radar).

All models were implemented in PyTorch \citep{paszke2019pytorch} and trained on NVIDIA T4 GPUs. Reproducibility is ensured through a global random seed initialization (\texttt{seed=42}) across all operations. Network weights are optimized using Riemannian Adam (\texttt{geoopt}) configured with a $10^{-3}$ learning rate, $10^{-4}$ weight decay, and a batch size of 32. Gradient clipping is capped at an $L_2$ norm of $2.0$. Training proceeds for a maximum of 100 epochs, governed by an early stopping mechanism (15-epoch patience) tied to a $20\%$ validation holdout to prevent overfitting.

\textbf{Numerical Safeguards:} Standard 32-bit eigendecomposition frequently triggers \texttt{LinAlgError} exceptions on rank-deficient kinematic covariances. To ensure stable automatic differentiation through our nested mappings, we introduce a numerical failsafe. Input matrices are symmetrized ($\mathbf{X} = \frac{1}{2}(\mathbf{X} + \mathbf{X}^\top)$) and upcast to 64-bit precision. Crucially, to prevent gradient degeneration during pullback computations, a microscopic, uniformly distributed tie-breaking vector ($\sim 10^{-8}$) is injected into the matrix diagonal prior to decomposition, ensuring isolated eigenvalues and well-conditioned Fréchet derivatives.

\paragraph{Native Riemannian Classification (Track A).} 
For intrinsic classification, we replace standard Euclidean linear layers with the generalized Riemannian Multinomial Logistic Regression (RMLR) \citep{chen2024rmlr}, equipped with the target manifold's metric. RMLR constructs geodesic hyperplanes directly on the manifold. We benchmark our proposed metrics, NHM (using Hyperbolic RMLR) and NSM (using Spherical RMLR), against the popular Log-Euclidean Metric (LEM, flat curvature) and the Affine-Invariant Riemannian Metric (AIRM, non-positive curvature, using AIRM RMLR). 

\paragraph{Vectorized Bounded Representation (Track B).} 
For pipelines constrained to Euclidean deep learning layers, we implement the Rational Conformal Metric. The metric output is projected to a Euclidean vector, normalized via \texttt{BatchNorm1d}, and classified by a standard dense linear classifier. The scale parameter $\alpha \in \mathbb{R}^+$ is optimized as a learnable parameter initialized via a softplus activation ($\alpha = \text{softplus}(\alpha_{raw}) + 10^{-4}$).

Furthermore, to measure the intrinsic geometry of our datasets without introducing metric bias, we project the matrices to the Log-Euclidean tangent space and normalize the pairwise distances $d \in [0, 1]$ to prevent Optimal Transport solver divergence. We then construct a $k$-Nearest Neighbor graph ($k=5$) and compute the discrete Ollivier-Ricci Curvature (ORC) via random walk diffusion ($\alpha = 0.5$).

\subsection{Validation via Synthetic Geometric Control}
\label{subsec:synthetic_generation}
To validate the classification bounds of RMLR equipped with the Nested Spherical Metric (NSM), we synthesize a covariance dataset localized to a compact, positively curved submanifold. Generating data with explicit global positive curvature in high-dimensional SPD spaces requires structural construction, as standard ambient matrix distributions do not natively guarantee these geometric properties.

We achieve this by embedding a 3-dimensional spherical data distribution directly into the ambient space of symmetric matrices $\mathcal{S}^{20}$. We define three class clusters using spherical coordinates $(\theta, \phi)$ scaled by a radial parameter $r$. These coordinates are embedded into the top-left $2 \times 2$ principal submatrix, padded with low-magnitude isotropic noise ($\mathcal{N}(0, 0.1)$) across the remaining dimensions, and mapped onto the SPD manifold via the matrix exponential. 

By modulating the radial parameter $r$, we dictate the prominence of the spherical manifold structure relative to the ambient noise, synthesizing distributions of opposing intrinsic curvatures:
\begin{itemize}
    \item \textbf{Macroscopic ($r=1.5$):} The spherical signal dominates the ambient noise ($\mathcal{N}(0, 0.1)$), forming distinct clusters separated by angular great-circle arcs along the curved shell. This empirically induces positive intrinsic curvature (Test ORC: $+0.010$). The Nested Spherical Metric (NSM) natively matches this geometry. 
    
    \item \textbf{Microscopic ($r=0.25$):} The spherical signal collapses into the noise floor, producing a dense Gaussian cluster near the origin. This shift induces strongly negative intrinsic curvature (Test ORC: $-0.213$). In this negatively curved geometry, NSM's sub-Euclidean volume growth artificially crowds the space and fails to resolve the margins. Conversely, NHM's super-Euclidean volume growth aligns with the negative intrinsic curvature, enhancing class separability and restoring accuracy.
\end{itemize}

\subsection{Geometric Regularization against Adversarial Outliers}
\label{subsec:adversarial_generation}
To empirically validate the Tangent Space Mapping (TSM) framework under the geodesically incomplete Rational Conformal Metric (Track B), we evaluate its robustness against spatially extreme anomalies. Because the standard Log-Euclidean tangent space is unbounded, large-magnitude outliers incur large penalties under cross-entropy optimization. This forces linear decision boundaries to shift, systematically degrading the separability of the core data distribution.

We define four linearly separable core classes (200 total samples) centered in the four quadrants of a 2D Euclidean feature space. To evaluate geometric robustness, we inject strategic spatial anomalies designed to disrupt linear separability:
\begin{enumerate}
    \item \textbf{Cross-Boundary Outlier Injection:} Anomalies belonging to Class 3 are embedded deep within the spatial domain of Class 0.
    \item \textbf{Directional Heavy-Tailed Noise:} A high-variance structural noise tail from Class 1 extends linearly across the origin, intersecting opposing class distributions.
    \item \textbf{Symmetric Extreme Outliers:} Distant, high-density point masses for Classes 0 and 3 are placed at opposite spatial extremes, generating high-magnitude gradient signals that disproportionately dictate the hyperplane optimization.
\end{enumerate}

Under standard optimization (Adam, lr=$0.1$, 600 epochs to ensure convergence), the unbounded Log-Euclidean baseline misclassifies the dense core clusters to minimize the disproportionate cross-entropy loss incurred by the outliers. Conversely, mapping these identical datasets through the Rational conformal transformation applies a non-linear radial compression. By asymptotically confining all representations within the open ball of radius $1/\alpha$, the spatial magnitude and corresponding gradient contribution of the anomalies are bounded. This geometric regularization isolates the linear optimization to the core distributions, recovering $100\%$ accuracy when using RCM without explicit outlier filtering.

\section{Results}
\label{sec:experiments}

A fundamental premise of our work is that metric curvature must align with data geometry. Our ORC analyses reveal negative intrinsic curvature across real-world datasets used here. The First-Person Hand Action (FPHA) kinematic dataset yields negative values (Train ORC: $-0.0969$, Test ORC: $-0.1197$), indicative of a hierarchical tree structure, consistent with the articulated kinematic tree structure governing human hand movements. Similarly, the Radar covariance dataset exhibits negative curvature (Train ORC: $-0.1780$, Test ORC: $-0.0646$). These results predict that imposing positive spherical curvature will induce geometric overlapping.

\subsection{Main Benchmark Results (Track A and Track B)}
We benchmark our nested inductive biases across a shallow Linear Probe, SPDNet, and a deep Riemannian ResNet (RResNet). Geodesically complete spaces are evaluated via native RMLR (Track A). The geodesically incomplete Rational Conformal Metric is evaluated via Tangent Space Mapping (Track B). Standard Log-Euclidean representations are evaluated under both frameworks to serve as the baseline.

\begin{table}[h]
\centering
\caption{Native Riemannian Classification (RMLR) on the FPHA Dataset.}
\label{tab:rmlr_fpha}
\begin{tabular}{lcccc}
\toprule
\textbf{Architecture} & \textbf{LEM (Flat)} & \textbf{AIRM} & \textbf{NHM (Hyperbolic)} & \textbf{NSM (Spherical)} \\
\midrule
Linear Probe & 0.8539 & 0.8417 & \textbf{0.8557} & 0.1443 \\
SPDNet       & 0.8191 & \textbf{0.8626} & 0.8504 & 0.1374 \\
RResNet      & 0.7652 & 0.7930 & \textbf{0.8643} & 0.0435 \\
\bottomrule
\end{tabular}
\end{table}

\begin{table}[h]
\centering
\caption{Native Riemannian Classification (RMLR) on the Radar Dataset.}
\label{tab:rmlr_radar}
\begin{tabular}{lcccc}
\toprule
\textbf{Architecture} & \textbf{LEM (Flat)} & \textbf{AIRM} & \textbf{NHM (Hyperbolic)} & \textbf{NSM (Spherical)} \\
\midrule
Linear Probe & 0.9600 $\pm$ 0.006 & 0.9513 $\pm$ 0.005 & \textbf{0.9633 $\pm$ 0.009} & 0.8163 $\pm$ 0.019 \\
SPDNet       & 0.9613 $\pm$ 0.008 & 0.9520 $\pm$ 0.008 & \textbf{0.9633 $\pm$ 0.009} & 0.7993 $\pm$ 0.029 \\
RResNet      & 0.9590 $\pm$ 0.012 & 0.9507 $\pm$ 0.006 & \textbf{0.9607 $\pm$ 0.011} & 0.5893 $\pm$ 0.121 \\
\bottomrule
\end{tabular}
\end{table}

\begin{table}[h]
\centering
\caption{Vectorized Tangent Space Mapping (TSM) on the FPHA Dataset.}
\label{tab:tsm_fpha}
\begin{tabular}{lcc}
\toprule
\textbf{Architecture} & \textbf{LEM (Flat Baseline)} & \textbf{RCM (Proposed)} \\
\midrule
Linear Probe & 0.8539 & \textbf{0.8730} \\
SPDNet       & \textbf{0.8748} & 0.8678 \\
RResNet      & \textbf{0.8852} & 0.8817 \\
\bottomrule
\end{tabular}
\end{table}

\begin{table}[h]
\centering
\caption{Vectorized Tangent Space Mapping (TSM) on the Radar Dataset.}
\label{tab:tsm_radar}
\begin{tabular}{lcc}
\toprule
\textbf{Architecture} & \textbf{LEM (Flat Baseline)} & \textbf{RCM (Proposed)} \\
\midrule
Linear Probe & 0.9617 $\pm$ 0.005 & \textbf{0.9640 $\pm$ 0.007} \\
SPDNet       & 0.9590 $\pm$ 0.008 & \textbf{0.9617 $\pm$ 0.007} \\
RResNet      & \textbf{0.9623 $\pm$ 0.006} & 0.9513 $\pm$ 0.017 \\
\bottomrule
\end{tabular}
\end{table}
As detailed in Tables \ref{tab:rmlr_fpha} and \ref{tab:rmlr_radar}, the native RMLR results (Track A) strictly corroborate our intrinsic dataset curvature findings. The nested Hyperbolic metric (NHM) consistently shows superior performance on the intrinsically negatively curved datasets. Most notably, in the deep FPHA RResNet architecture, NHM achieves a prominent +9.9\% absolute accuracy gain over the baseline Log-Euclidean RMLR and a +7.1\% gain over the standard AIRM benchmark. Conversely, the Spherical metric (NSM) exhibits substantial margin collapse, demonstrating the need for metric alignment to preserve separability.

In Track B, designed for pipelines reliant on standard Euclidean dense layers, our RCM Tangent Space Mapping (TSM) (Tables \ref{tab:tsm_fpha} and \ref{tab:tsm_radar}) achieves competitive performance. While RCM's strict spatial bounding induces a minor regularization penalty on clean datasets within deeper architectures, this bounding is precisely what enables its state-of-the-art robustness against severe adversarial anomalies. Its true superiority is realized in its geometric regularization against outliers, as detailed in Section \ref{subsec:rational_warp_results}.

\subsection{Validation via Synthetic Geometric Control Results}
\label{subsec:synthetic_ablation}
To ensure the performance degradation of the Nested Spherical Metric (NSM) on real-world data was a consequence of intrinsic geometric mismatch rather than algorithmic instability, we evaluated the RResNet architecture on our controlled synthetic dataset (Section \ref{subsec:synthetic_generation}). By modulating the radial parameter $r$, we shift the data distribution between positive and negative intrinsic curvature regimes (test sets), measured via the discrete Ollivier-Ricci Curvature (ORC).

\begin{table}[h]
\centering
\caption{Classification Test Set Accuracy on Synthetic Geometric Control Dataset.}
\label{tab:synthetic_ablation}
\begin{tabular}{lccc}
\toprule
\textbf{Radial Scale} & \textbf{Intrinsic Curvature (Test ORC)} & \textbf{NHM} & \textbf{NSM} \\
\midrule
Macroscopic ($r=1.5$) & $+0.010$ & $0.9950$ & $\mathbf{1.0000}$ \\
Microscopic ($r=0.25$) & $-0.213$ & $\mathbf{0.8000}$ & $0.6875$ \\
\bottomrule
\end{tabular}
\end{table}

As detailed in Table \ref{tab:synthetic_ablation}, the empirical results validate the volume-growth mechanics established in our experimental setup. At the macroscopic scale ($r=1.5$, positive ORC), NSM natively matches the spherical geometry to achieve perfect accuracy ($1.0000$), while NHM's radial distance distortion slightly degrades separability ($0.9950$). 

However, at the microscopic scale ($r=0.25$, negative ORC), the sub-Euclidean volume growth of NSM fails to match the negatively curved noise-dominated core, degrading performance ($0.6875$). NHM aligns with the negative curvature; its super-Euclidean expansion disperses the overlapping representations, pulling the classes apart and restoring accuracy to $0.8000$. This symmetrical ablation demonstrates that imposing Riemannian priors is only effective when explicitly aligned with the dataset's intrinsic geometry.

\subsection{Geometric Regularization: Outlier Results (Track B)}
\label{subsec:rational_warp_results}

\begin{figure}[h]
    \centering
    \includegraphics[width=\textwidth]{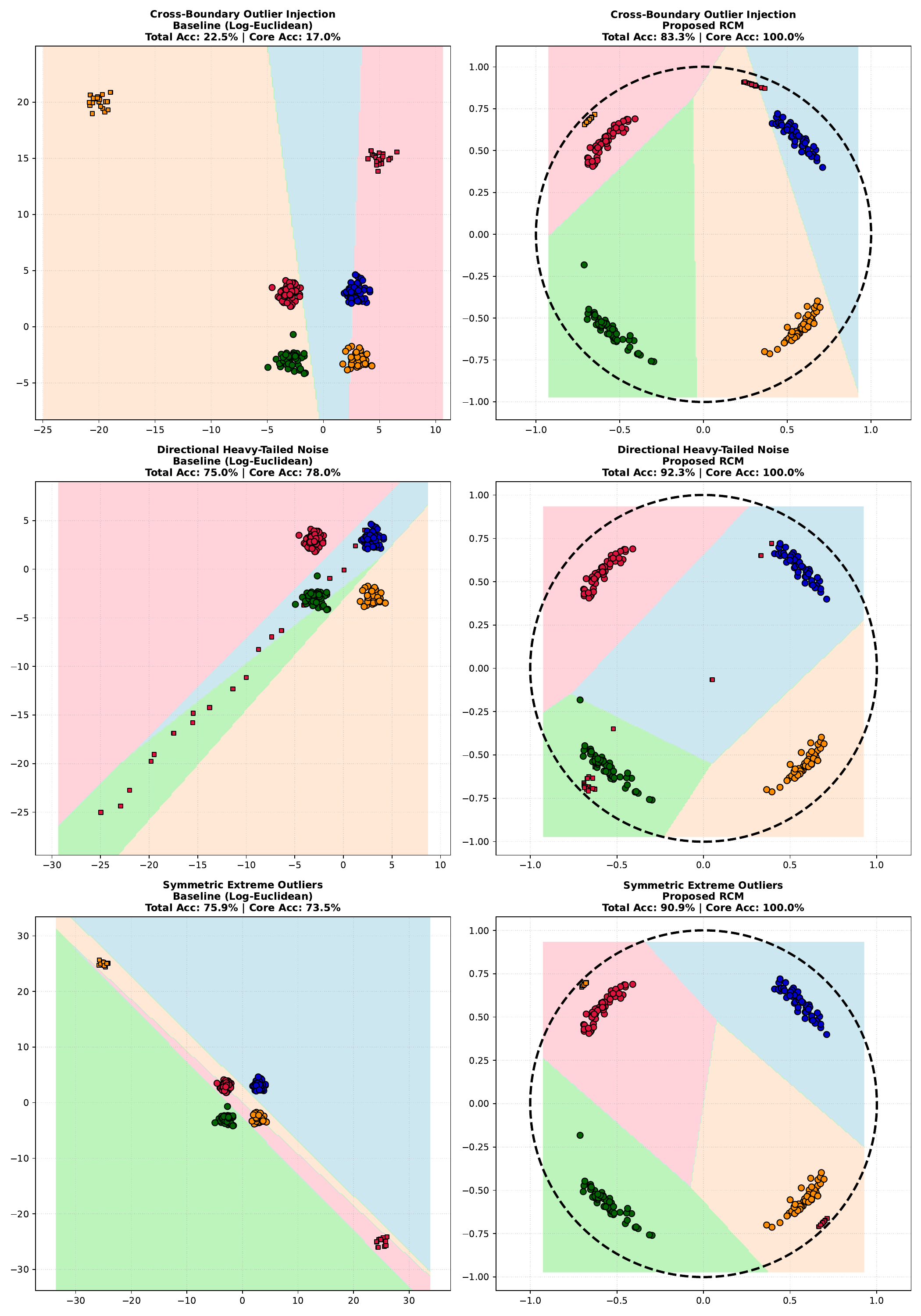}
    \caption{Decision boundaries under adversarial spatial anomalies across three topologies: cross-boundary outlier injection (top), directional heavy-tailed noise (middle), and symmetric extreme outliers (bottom). The unbounded Log-Euclidean baseline (left) systematically shifts hyperplanes to minimize outlier cross-entropy penalties, degrading core separability. The proposed Rational Conformal Metric (right) asymptotically confines outliers within a fixed radial envelope (dashed circle), mathematically bounding their gradient contribution and restoring perfect core accuracy.}
    \label{fig:adversarial_scenarios}
\end{figure}

\begin{table}[h]
\centering
\caption{Classification Accuracy under Adversarial Spatial Outliers. The unbounded Log-Euclidean baseline sacrifices core separability to accommodate extreme anomalies, while the proposed Rational Conformal Metric (RCM) mathematically bounds their gradient contribution, restoring perfect core accuracy.}
\label{tab:adversarial_results}
\begin{tabular}{lcccc}
\toprule
 & \multicolumn{2}{c}{\textbf{Baseline (Log-Euclidean)}} & \multicolumn{2}{c}{\textbf{Proposed (RCM)}} \\
\cmidrule(lr){2-3} \cmidrule(lr){4-5}
\textbf{Adversarial Experiments} & \textbf{Total Acc.} & \textbf{Core Acc.} & \textbf{Total Acc.} & \textbf{Core Acc.} \\
\midrule
Cross-Boundary Outlier Injection & $22.5\%$ & $17.0\%$ & $83.3\%$ & $\mathbf{100.0\%}$ \\
Directional Heavy-Tailed Noise & $75.0\%$ & $78.0\%$ & $\mathbf{92.3\%}$ & $\mathbf{100.0\%}$ \\
Symmetric Extreme Outliers & $75.9\%$ & $73.5\%$ & $90.9\%$ & $\mathbf{100.0\%}$ \\
\bottomrule
\end{tabular}
\end{table}
As illustrated in Figure \ref{fig:adversarial_scenarios} (Appendix \ref{app:visualizations}) and quantitatively detailed in Table \ref{tab:adversarial_results}, the unbounded Log-Euclidean baseline is highly susceptible to unconstrained spatial anomalies. Because outliers incur large cross-entropy penalties, the linear optimizer systematically shifts the decision hyperplanes to compensate. This global hyperplane shift degrades the separability of the core data distributions, dropping core accuracy to $22.5\%$ under cross-boundary outlier injection.

However, mapping the identical feature spaces through the Rational conformal metric transformation ($\mathbf{v} \cdot (1 + \alpha \|\mathbf{v}\|_2)^{-1}$) asymptotically confines all representations within an open Euclidean ball of radius $1/\alpha$. By bounding the spatial magnitude of the outliers, their gradient contribution during backpropagation is limited. The linear classifier is subsequently able to optimize the hyperplanes relative to the dense core distributions, stabilizing the decision boundaries and restoring core classification accuracy to state-of-the-art levels across all adversarial variants.

\begin{figure}[h]
    \centering
    \includegraphics[width=\textwidth]{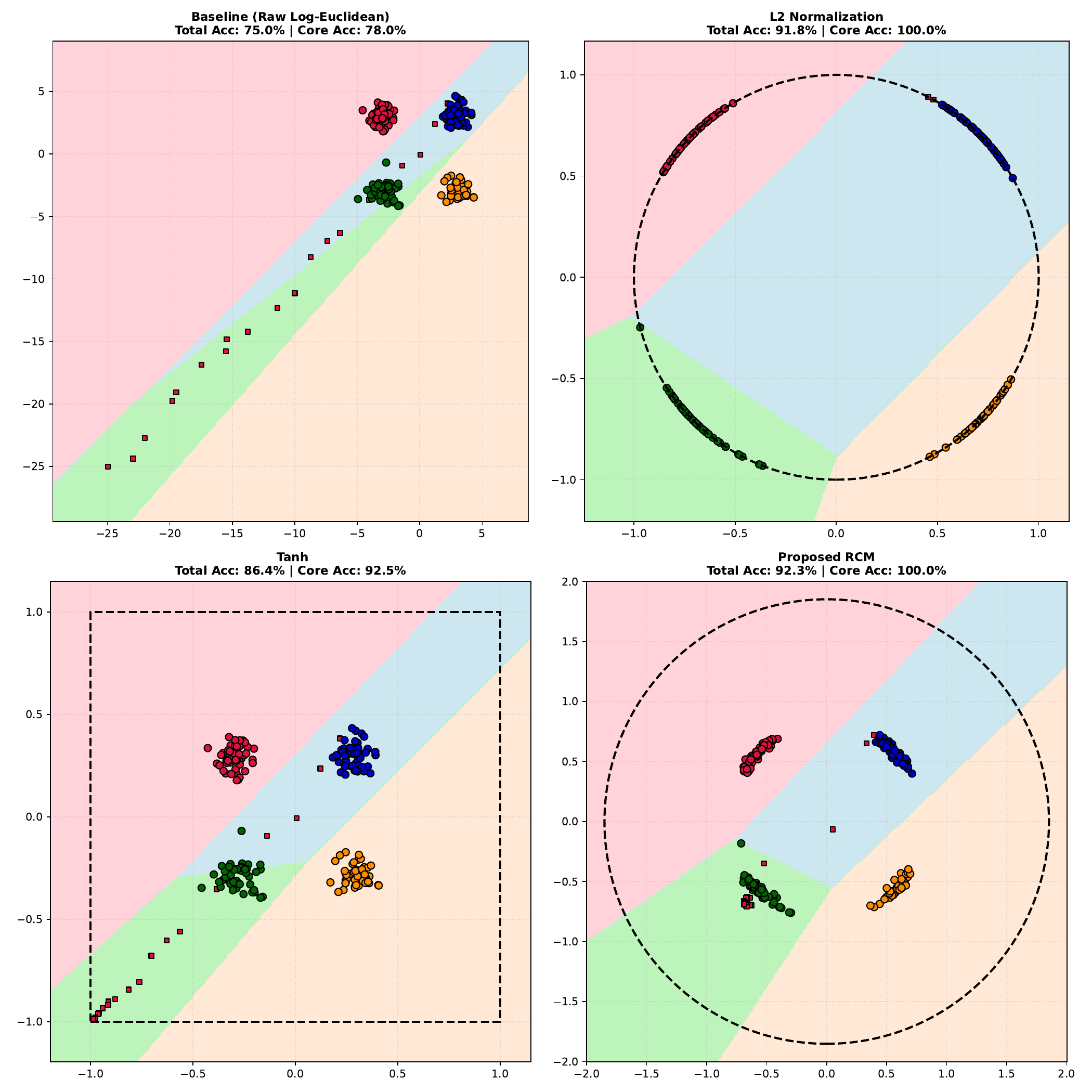}
    \caption{Ablation of geometric regularization techniques on the directional heavy-tailed distribution. While $L_2$ normalization (top right) restores core accuracy, it collapses the 2D Gaussian distributions into rigid 1D arcs, discarding intrinsic data magnitude, leading to lesser total accuracy than RCM. Tanh squashing (bottom left) bounds the space but introduces anisotropic box distortion, skewing inter-class angles. The proposed Rational Conformal Metric (bottom right) optimally bounds outlier magnitude while preserving the relative spatial variance and angular separation of the primary data.}
    \label{fig:adversarial_ablation}
\end{figure}

\textbf{Ablation:} To isolate the structural advantages of this specific radial compression, we ablated the Rational mapping against standard Euclidean normalization techniques using the directional heavy-tailed distribution (Figure \ref{fig:adversarial_ablation}, Appendix \ref{app:visualizations}). 
\begin{itemize}
    \item \textbf{$L_2$ Normalization ($\mathbf{v}/\|\mathbf{v}\|_2$):} While $L_2$ projection mathematically bounds outlier gradients and successfully restores core separability (Core Accuracy: $100.0\%$, Total Accuracy: $91.8\%$), it eliminates intrinsic data magnitude. This projects the 2D Gaussian distributions onto fixed-radius 1D arcs, discarding the latent variance of the primary data.
    \item \textbf{Tanh ($\tanh(\beta \mathbf{v})$):} Applying an element-wise hyperbolic tangent bounds the vector space but introduces anisotropic distortion by compressing the space toward a hypercube. This non-uniform scaling skews inter-class angles, preventing the recovery of optimal linear margins (Core Accuracy: $92.5\%$, Total Accuracy: $86.4\%$).
    \item \textbf{Proposed RCM:} The proposed mapping provides the optimal geometric compromise. It enforces the mathematical bounds necessary to limit outlier impact while preserving the relative spatial variance and angular separation of the core distributions near the origin. This yields superior global performance (Core Accuracy: $100.0\%$, Total Accuracy: $92.3\%$) without inducing dimensional reduction or angular distortion.
\end{itemize}
This establishes the Rational conformal mapping as a uniquely robust geometric prior for mitigating unbounded sensor noise in deep learning architectures.

\section{Conclusion}
\label{sec:conclusion}
In this work, we introduced a nested inductive bias framework for Symmetric Positive Definite (SPD) manifolds, bridging the theoretical gap between native matrix geometries and established vector-space geometric priors. By utilizing nested diffeomorphisms, we pulled back Hyperbolic and Spherical geometries to construct curvature-aligned Riemannian classifiers. Our empirical evaluations and synthetic ablations demonstrate that deep manifold networks experience significant degradation in class separability unless the metric curvature aligns with the intrinsic geometry of the data distribution. Furthermore, for standard vectorized architectures, our proposed Rational Conformal Metric (RCM) imposes a radial bound on the representation space, yielding state-of-the-art robustness against extreme spatial anomalies. A direction for future research involves extending this pullback framework beyond the terminal classification layer to construct fully nested Riemannian dense layers, enabling the end-to-end preservation of specific non-Euclidean curvatures across all hidden representations.

\clearpage
\bibliography{iclr2027_conference}
\bibliographystyle{iclr2027_conference}
\clearpage
\appendix
\section{Appendix}
\subsection{Adversarial Outlier Regularization Visualizations}
\label{app:visualizations}

This section provides the decision boundary visualizations for the geometric regularization experiments.

\end{document}